\documentclass{article}
\usepackage{ijcai21}

\usepackage{times}
\usepackage{soul}
\usepackage{url}
\usepackage[hidelinks]{hyperref}
\usepackage[utf8]{inputenc}
\usepackage[small]{caption}
\usepackage{graphicx}
\usepackage{amsmath}
\usepackage{amsthm}
\usepackage{booktabs}
\usepackage{caption} 
\usepackage{amsmath,amssymb} 
\usepackage{multirow} 
\usepackage[normalem]{ulem}
\useunder{\uline}{\ul}{}

\usepackage{booktabs}
\usepackage{xspace}

\newcommand{\bW}{\mathbf W}
\let\oldnl\nl
\newcommand{\nonl}{\renewcommand{\nl}{\let\nl\oldnl}}

\def\eg{\emph{e.g., }} 

\def\ie{\emph{i.e., }} 

\def\etc{\emph{, etc.}}

\relax

\title{Chop Chop BERT: Visual Question Answering by Chopping VisualBERT’s Heads}

\author{
Chenyu Gao$^{1,2,3}$
\and
Qi Zhu$^1$\and
Peng Wang$^{1,3}$ \footnote{Peng Wang is corresponding author.}\And
Qi Wu$^4$
\affiliations
$^1$School of Computer Science, Northwestern Polytechnical University, Xi'an, China\\
$^2$School of Software, Northwestern Polytechnical University, Xi'an, China\\
$^3$National Engineering Laboratory for Integrated Aero-Space-Ground-Ocean Big Data Application Technology, China\\
$^4$University of Adelaide, Australia
\emails
\{chenyugao, zhu$\_$qi$\_$happy$\_$\}@mail.nwpu.edu.cn, peng.wang@nwpu.edu.cn, qi.wu01@adelaide.edu.au
}

\begin{document}

\maketitle

\begin{abstract}
Vision-and-Language (VL) pre-training has shown great potential on many related downstream tasks, such as Visual Question Answering (VQA), one of the most popular problems in the VL field. All of these pre-trained models (such as VisualBERT, ViLBERT, LXMERT and UNITER) are built with Transformer, which extends the classical attention mechanism to multiple layers and heads. To investigate why and how these models work on VQA so well, in this paper we explore the roles of individual heads and layers in Transformer models when handling $12$ different types of questions. Specifically, we manually remove (chop) heads (or layers) from a pre-trained VisualBERT model at a time, and test it on different levels of questions to record its performance. As shown in the interesting echelon shape of the result matrices, experiments reveal different heads and layers are responsible for different question types, with higher-level layers activated by higher-level visual reasoning questions. Based on this observation, we design a dynamic chopping module that can automatically remove heads and layers of the VisualBERT at an instance level when dealing with different questions. Our dynamic chopping module can effectively reduce the parameters of the original model by $50\%$, while only damaging the accuracy by less than $1\%$ on the VQA task.

\end{abstract}

\section{Introduction}
Transformer~\cite{vaswani2017attention} architecture was designed for translation tasks but has shown good performance across many other tasks, ranging from text summarization~\cite{egonmwan2019transformer,xu2020discourse}, language modeling~\cite{dai2019transformer,ma2019tensorized} to question answering~\cite{shao2019transformer,sanh2019distilbert}.  The BERT (Bidirectional Encoder Representations from Transformers)~\cite{devlin2018bert} model is then proposed for general pre-training. By fine-tuning the pre-trained model on downstream tasks, BERT further advances the state-of-the-art for multiple NLP tasks.
Inspired by it, many variants have been proposed towards the solution of vision-and-language tasks, \eg Visual Question Answering (VQA~\cite{kafle2017visual}), including VisualBERT~\cite{li2019visualbert}, VL-BERT~\cite{su2019vl}, ViLBERT~\cite{lu2019vilbert}, LXMERT \cite{tan2019lxmert}, and UNITER \cite{chen2019uniter}\etc

These vision-language BERT models are all Transformer-based, where the most innovative part is the multi-layer and multi-head settings, compared to the classic attention. However, how the heads and layers contribute to those reasoning-required tasks such as Visual Question Answering, remains a mystery. 
Here we want to explore the role of heads and layers and focus on the VQA task, hoping to find their patterns that human can `see' and `understand'. When faced with high-level tasks, there are different areas activated in our brain to fulfill different tasks. Our findings reveal that different heads and layers are indeed activated by different question types.

For the first time, we come up with a statistical analysis of Transformer's heads and layers triggered off by $12$ types of questions on the Task Driven Image Understanding Challenge (TDIUC)~\cite{kafle2017analysis}. The TDIUC dataset is a large VQA dataset with $12$ more fine-grained categories proposed to compensate for the bias in distribution of different question types of VQA 2.0~\cite{goyal2017making}, which provide convenience for our analysis. Our experiments based on the VisualBERT, as for it's general Transformer style architecture without more extra designs. The experiments can be divided into two parts: Manual Chopping and Dynamic Chopping. 
For manual chopping in Section \ref{sec:effect_manual_chopping}, we explored the effect of a particular head by removing it, or removing all the other heads in a layer but that one. As the most obvious pattern is found at a layer level, we further experiment by chopping one layer at a time. As there are $12$ kinds of questions in TDIUC and $12$ layers in VisualBERT, we get a result matrix with the size $12 \times 12$. The echelon shape in this matrix reveal different layers play different roles across $12$ types of question, with higher-level reasoning questions demanding higher-level Transformer layers gradually. For Dynamic Chopping, we chop the layers simultaneously ( Section \ref{sec:layre_chopping}) under the control of instance-conditioned layer scores, which are generated by a dynamic chopping module. We further find that removing some layers can decrease parameters but increase the accuracy of models.

Overall, the major contribution of this work is to delve into the roles of Transformer heads and layers in different types of visual demanding questions for the first time. We hope our interesting finding, especially the presented echelon shape of experiment results matrices, which shows a clear tendency of Transformer to gradually rely on higher-level layers when reaching higher-level reasoning questions, can inspire further investigation of the inner mechanism of self-attention layers.

\section{Related Work $\&$ Background}
\subsection{Transformer in VQA}
Transformer-based models, such as VisualBERT~\cite{li2019visualbert} and ViLBERT~\cite{lu2019vilbert}, have gained popularity most recently, as they are well suited to pre-training and can be easily transferred to other similar tasks.
VisualBERT consists of a stack of Transformer layers that implicitly align elements of an input text and regions in an associated image with self-attention.
ViLBERT proposes to learn joint representation of images and text using a BERT-like architecture with separate Transformers for vision and language that can attend to each other.
LXMERT~\cite{tan2019lxmert} is a large-scale Transformer model that consists of three encoders: an object relationship encoder, a language encoder, and a cross-modality encoder. 
UNITER~\cite{chen2019uniter} adopts Transformer to leverage its self-attention mechanism designed for learning contextualized representations.

\subsection{Multi-head $\&$ Multi-layer Transformer}
Transformer~\cite{vaswani2017attention} architecture relies entirely on the self-attention mechanism. The self-attention $\textbf{Attn}$ is calculated with a set of queries $\{q_i\}_{i=1}^{N}$, a set of keys $\{k_i\}_{i=1}^{N}$ and a set of values $\{v_i\}_{i=1}^{N}$:
\begin{equation}
{
	\small{\begin{split}
    	& \mathbf{attn_i} = \sum_{j=1}^{N}\mathbf{softmax}({q_ik_j} / {\sqrt{d}})v_j, \\
    	& \mathbf{Attn} = [attn_1, attn_2, ..., attn_N]^T,
	\end{split}
	\label{layer_score}}}
\end{equation}
where $N$ is the number of key-value pairs, $d$ is the dimension of keys and values, and ${1} / {\sqrt{d}}$ is a scaling factor.

With a total of $H$ heads, multi-head attention can be regarded as an ensemble of each head $h$'s self-attention:
\begin{equation}
{
	\begin{split}
    	& \mathbf{{MH}\_{Attn}} = \mathbf{Concat}(\mathbf{Attn}_1, ..., \mathbf{Attn}_h, ..., \mathbf{Attn}_H),
	\end{split}
	\label{layer_score}
}
\end{equation}
where $H=12$ in our paper. 

A Transformer encoder contains a total of $L$ layers, with each one including two sub-layers: a multi-head self-attention layer and a fully connected feed forward layer. All of the queries and key-value pairs in a same self-attention layer come from the output of the previous layer in the encoder. Therefore, each position in a layer of encoder can attend to all the positions in the previous layer. In our paper, we set $L=12$.

For each layer, there is a multi-head attention layer. The multi-head attention in layer $l$ can be written as:
\begin{equation}
{
	\begin{split}
    	& \mathbf{{MH}\_{Attn}}^l = \mathbf{Concat}(\mathbf{Attn}^l_1, ..., \mathbf{Attn}^l_h, ..., \mathbf{Attn}^l_H),
	\end{split}
	\label{layer_score}
}
\end{equation}
where $l$ ranges from $1$ to $L$.

\section{Chopping Method}
We perform chopping upon our base model -- VisualBERT, which is a general Transformer style architecture without extra designs. 
Viewing one head as an element, we get an $L \times H = 12 \times 12$ grid-like structure for the model. 
The purpose of chopping is to disable a part of the model and make it unable to participate in the forward process. For heads and layers, we have separate strategies to reach this end, as they have different characteristics in cooperating. Heads work in parallel while layers work in a sequential manner.
\subsection{Manual Chopping}
In manual chopping, whether to chop a component or not is controlled by our signal manually. The signals are given at two levels: heads and layers.
\subsubsection{Manual Chopping of Heads}
\begin{figure}[t!]
	\begin{center}
		\includegraphics[width=0.48\textwidth]{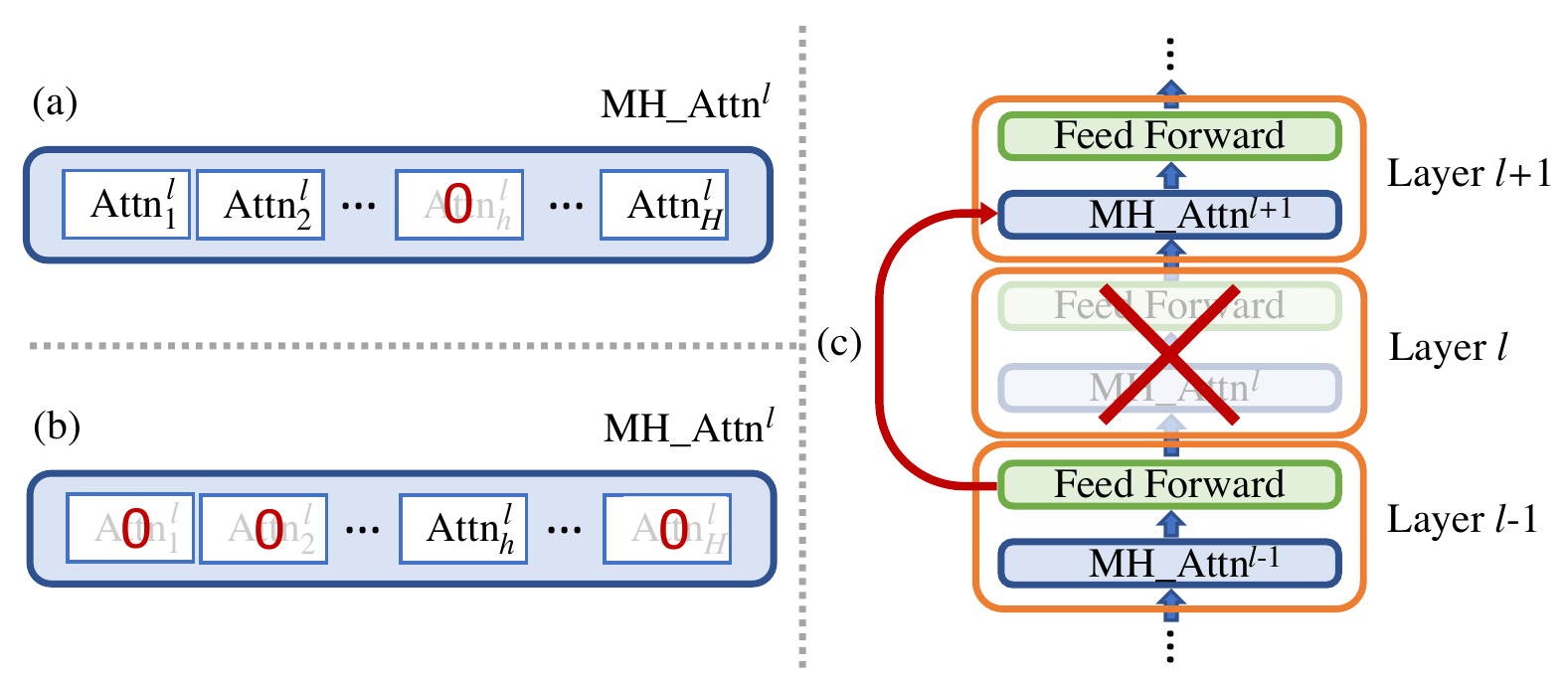}
	\end{center}
	\caption{(a) to remove one head $h$ in layer $l$, the self-attention weights $\textbf{Attn}_h^l$ are multiplied with zero-valued binary mask. (b) to study whether one head $h$ in layer $l$ can replace the whole layer, all of the self-attention weights in layer $l$ are multiplied with zero-valued binary mask except $\textbf{Attn}_h^l$. (c) A part of Transformer encoder layers. After chopping Layer $l$, the output of Layer $l-1$ becomes the input of Layer $l+1$. }
	\label{fig:chop_layer}
\end{figure}

The importance of each head is investigated thoroughly. Every time, We simply modified the multi-head attention in a specific layer $l$ by multiplying one head attention by a binary mask variable $\alpha^l_h$ ($h = 1, 2, ..., H$): 
\begin{equation}
{
	\begin{split}
    	& \mathbf{{MH}\_{Attn}}^l = \mathbf{Concat}(\alpha^l_1\mathbf{Attn}^l_1, ...,  \alpha^l_2\mathbf{Attn}^l_h, ..., \alpha^l_H\mathbf{Attn}^l_H),
	\end{split}
	\label{layer_score}}
\end{equation}
where $\alpha^l_h$ can be either $0$ or $1$. The Transformer model we used have $12$ layers with $12$ heads in each layer, there are totally $144$ $\alpha^l_h$. For each time of removing one head, only one $\alpha$ is set to $0$ and all the others are set to $1$ (in Figure \ref{fig:chop_layer} (a)). As for removing all heads but one, only one $\alpha$ is set to $1$ and all the others are set to $0$ in a specific layer (in Figure \ref{fig:chop_layer} (b)). To study the contribution of each head, we exhaust each $\alpha^l_h$ in the above two experiments.

\subsubsection{Manual Chopping of Layers}
As the layers are in a serial manner, to chop off one specific Layer $l$, we directly skip it and connect the output of Layer $l-1$ to the input of Layer $l+1$. The detail is shown in Figure \ref{fig:chop_layer} (c). We chop each layer in the Transformer one time.

\begin{center}
	\begin{table*}[t]
	\centering
	\scalebox{0.9}{
	\setlength{\tabcolsep}{0.9mm}{
		\begin{tabular}{clcccccccccccc}
        \toprule[1pt]
        \# &
          \multicolumn{1}{c}{Question types} &
          L1 &
          L2 &
          L3 &
          L4 &
          L5 &
          L6 &
          L7 &
          L8 &
          L9 &
          L10 &
          L11 &
          L12 \\ \hline\hline
        1 &
          Absurd &
          $-0.18\%$ &
          $-0.75\%$ &
          $-0.56\%$ &
          {\ul $\textbf{-5.73\%}$} &
          $0.23\%$ &
          $-0.13\%$ &
          $0.05\%$ &
          $0.02\%$ &
          $-0.03\%$ &
          $0.04\%$ &
          $-0.05\%$ &
          $-0.03\%$ \\
        2 &
          Obj Pres &
          $-0.12\%$ &
          $-0.79\%$ &
          $-0.05\%$ &
          $-0.33\%$ &
          $-0.18\%$ &
          $0.08\%$ &
          $0.01\%$ &
          $0.02\%$ &
          $0.03\%$ &
          $0.03\%$ &
          $0.02\%$ &
          $-0.02\%$ \\
        3 &
          Scene Rec &
          $-0.22\%$ &
          $0.19\%$ &
          $-0.36\%$ &
          $-0.19\%$ &
          $-0.19\%$ &
          $-0.12\%$ &
          $0.10\%$ &
          $-0.06\%$ &
          $0.11\%$ &
          $-0.07\%$ &
          $0.04\%$ &
          $-0.10\%$ \\
        4 &
          Sub-Obj Rec &
          $0.07\%$ &
          $-0.61\%$ &
          $-0.16\%$ &
          $-0.30\%$ &
          $-0.39\%$ &
          $0.06\%$ &
          $-0.05\%$ &
          $-0.05\%$ &
          $0.06\%$ &
          $0.05\%$ &
          $0.06\%$ &
          $0.09\%$ \\
        5 &
          Sport Rec &
          $-0.13\%$ &
          $-0.21\%$ &
          $0.09\%$ &
          $-0.36\%$ &
          $-0.20\%$ &
          $-0.03\%$ &
          $-0.10\%$ &
          $-0.06\%$ &
          $-0.06\%$ &
          $-0.09\%$ &
          $-0.05\%$ &
          $-0.05\%$ \\
        6 &
          Color Attr &
          $-0.25\%$ &
          {\ul $\textbf{-3.09\%}$} &
          -0.40\% &
          {\ul $\textbf{-1.26\%}$} &
          $-0.60\%$ &
          $-0.22\%$ &
          $-0.18\%$ &
          $-0.06\%$ &
          $0.06\%$ &
          $-0.07\%$ &
          $-0.07\%$ &
          $-0.13\%$ \\
        7 &
          Sentiment &
          $-0.48\%$ &
          $-0.73\%$ &
          {\ul $\textbf{-4.36\%}$} &
          $-0.97\%$ &
          $0.24\%$ &
          {\ul $\textbf{-1.45\%}$} &
          $0.24\%$ &
          $0.24\%$ &
          $0.24\%$ &
          $0.24\%$ &
          $0.24\%$ &
          $0.24\%$ \\
        8 &
          Count &
          {\ul $\textbf{-4.96\%}$} &
          {\ul $\textbf{-2.03\%}$} &
          {\ul $\textbf{-2.45\%}$} &
          {\ul $\textbf{-7.09\%}$} &
          {\ul $\textbf{-1.08\%}$} &
          $-0.34\%$ &
          $-0.28\%$ &
          $-0.27\%$ &
          $-0.41\%$ &
          $-0.25\%$ &
          $-0.13\%$ &
          $-0.08\%$ \\
        9 &
          Positional Rec &
          $0.33\%$ &
          {\ul $\textbf{-1.87\%}$} &
          $-0.54\%$ &
          {\ul $\textbf{-1.78\%}$} &
          {\ul $\textbf{-1.54\%}$} &
          $-0.57\%$ &
          $0.45\%$ &
          $0.57\%$ &
          $0.54\%$ &
          $0.63\%$ &
          $0.60\%$ &
          {\ul $\textbf{1.03\%}$} \\
        10 &
          Other Attr &
          $0.42\%$ &
          {\ul $\textbf{-3.58\%}$} &
          $0.63\%$ &
          {\ul $\textbf{-1.52\%}$} &
          {\ul $\textbf{-1.46\%}$} &
          {\ul $\textbf{-2.54\%}$} &
          $0.42\%$ &
          $0.17\%$ &
          $-0.66\%$ &
          $0.30\%$ &
          $0.36\%$ &
          $-0.36\%$ \\
        11 &
          Activity Rec &
          {\ul $\textbf{-1.12\%}$} &
          {\ul $\textbf{-3.13\%}$} &
          $0.88\%$ &
          {\ul $\textbf{-4.49\%}$} &
          {\ul $\textbf{1.04\%}$} &
          {\ul $\textbf{-5.46\%}$} &
          {\ul $\textbf{1.52\%}$} &
          {\ul $\textbf{2.01\%}$} &
          {\ul $\textbf{-1.04\%}$} &
          $0.56\%$ &
          $-0.80\%$ &
          {\ul $\textbf{-1.85\%}$} \\
        12 &
          Util \& Aff &
          {\ul $\textbf{-3.85\%}$} &
          {\ul $\textbf{-11.54\%}$} &
          {\ul $\textbf{1.92\%}$} &
          {\ul $\textbf{-5.77\%}$} &
          {\ul $\textbf{-9.62\%}$} &
          {\ul $\textbf{-5.77\%}$} &
          {\ul $\textbf{-5.77\%}$} &
          {\ul $\textbf{-5.77\%}$} &
          {\ul $\textbf{-1.92\%}$} &
          {\ul $\textbf{-1.92\%}$} &
          {\ul $\textbf{-1.92\%}$} &
          {\ul $\textbf{1.92\%}$} \\

        \bottomrule[1pt]
        \end{tabular}
	}
	}
	\caption{\label{tab:deltaAcc_removee_one}The relative difference of accuracy on each question type when only one head is removed, which is calculated by $(Acc_{new}-Acc_{org})/Acc_{org}$. In each layer, we only show the one with the \textbf{largest absolute value} out of $12$ heads. For the complete result of all $12$ heads, please see Appendix A. We conduct this head removal on TDIUC's $12$ question types. Underlined numbers indicate that its absolute value ($|(Acc_{new}-Acc_{org})/Acc_{org}|$) is above $1\%$. 
	}
	\end{table*}
\end{center}

\begin{center}
	\begin{table*}[t]
	\centering
	\scalebox{0.9}{
	\setlength{\tabcolsep}{0.9mm}{
		\begin{tabular}{clcccccccccccc}
        \toprule[1pt]
        \# &
          \multicolumn{1}{c}{Question types} &
          L1 &
          L2 &
          L3 &
          L4 &
          L5 &
          L6 &
          L7 &
          L8 &
          L9 &
          L10 &
          L11 &
          L12 \\ \hline\hline
        1  & Absurd      & $0.19\%$  & $0.60\%$  & $0.25\%$  & $-0.54\%$       & $0.88\%$        & $-0.01\%$      & $0.26\%$  & $0.09\%$ & $0.07\%$  & $0.17\%$  & $0.06\%$ & $0.07\%$  \\
        2 & Obj Pres    & $-0.51\%$ & $-0.49\%$ & $-0.21\%$ & {\ul $\textbf{-1.14\%}$} & $0.06\%$        & $0.10\%$       & $0.03\%$  & $0.10\%$ & $0.12\%$  & $0.08\%$  & $0.00\%$ & $-0.02\%$ \\
        3  & Scene Rec   & $-0.17\%$ & $0.09\%$  & $-0.17\%$ & $-0.38\%$       & $-0.14\%$       & $0.03\%$       & $0.27\%$  & $0.02\%$ & $0.14\%$  & $0.19\%$  & $0.19\%$ & $0.25\%$  \\
        4  & Sub-Obj Rec & $-0.10\%$ & $-0.34\%$ & $-0.49\%$ & $-0.39\%$       & {\ul $\textbf{-1.11\%}$} & 0.06\%       & $-0.17\%$ & $0.00\%$ & $-0.12\%$ & $0.12\%$  & $0.06\%$ & $0.10\%$  \\
        5  & Sport Rec   & $-0.35\%$ & $-0.30\%$ & $0.00\%$  & $-0.52\%$       & $-0.61\%$       & $-0.27\%$      & $-0.34\%$ & $0.01\%$ & $-0.02\%$ & $-0.12\%$ & $0.02\%$ & $0.07\%$  \\
        6 &
          Color Attr &
          {\ul $\textbf{-1.32\%}$} &
          {\ul $\textbf{-1.64\%}$} &
          {\ul $\textbf{-2.24\%}$} &
          {\ul $\textbf{-1.64\%}$} &
          {\ul $\textbf{-5.68\%}$} &
          $-0.68\%$ &
          $-0.61\%$ &
          $-0.15\%$ &
          $-0.01\%$ &
          $-0.20\%$ &
          $-0.05\%$ &
          $-0.09\%$ \\
        7 & Sentiment   & $0.73\%$  & $0.48\%$  & $0.24\%$  & $0.00\%$        & $0.48\%$        & {\ul $\textbf{1.21\%}$} & $0.00\%$  & $0.97\%$ & $0.97\%$  & $0.97\%$  & $0.24\%$ & $0.24\%$  \\ 
        8 &
          Count &
          {\ul $\textbf{-1.97\%}$} &
          {\ul $\textbf{-3.78\%}$} &
          {\ul $\textbf{-1.53\%}$} &
          {\ul $\textbf{-4.17\%}$} &
          {\ul $\textbf{-5.86\%}$} &
          {\ul $\textbf{-1.02\%}$} &
          $-0.66\%$ &
          $-0.40\%$ &
          $-0.59\%$ &
          $-0.30\%$ &
          $0.20\%$ &
          $-0.08\%$ \\
        9 &
          Positional Rec &
          $-0.99\%$ &
          {\ul $\textbf{-2.89\%}$} &
          {\ul $\textbf{-3.65\%}$} &
          {\ul $\textbf{-8.65\%}$} &
          {\ul $\textbf{-7.75\%}$} &
          {\ul $\textbf{-1.09\%}$} &
          {\ul $\textbf{-1.15\%}$} &
          $-0.27\%$ &
          $-0.03\%$ &
          $0.51\%$ &
          $0.93\%$ &
          $-0.81\%$ \\
        10 &
          Other Attr &
          $-0.49\%$ &
          $-0.97\%$ &
          {\ul $\textbf{-1.33\%}$} &
          $0.32\%$ &
          {\ul $\textbf{-4.68\%}$} &
          $-0.89\%$ &
          $0.21\%$ &
          $-0.30\%$ &
          $-0.99\%$ &
          $-0.40\%$ &
          $0.21\%$ &
          $-0.13\%$ \\
        11 &
          Activity Rec &
          $-0.08\%$ &
          $0.08\%$ &
          $-0.32\%$ &
          $-0.24\%$ &
          {\ul $\textbf{-2.65\%}$} &
          {\ul $\textbf{-2.25\%}$} &
          {\ul $\textbf{-2.41\%}$} &
          {\ul $\textbf{1.12\%}$} &
          {\ul $\textbf{-3.13\%}$} &
          {\ul $\textbf{-1.12\%}$} &
          $-0.40\%$ &
          $-0.56\%$ \\
        12 &
          Util \& Aff &
          $0.00\%$ &
          {\ul $\textbf{-9.62\%}$} &
          $0.00\%$ &
          {\ul $\textbf{-1.92\%}$} &
          {\ul $\textbf{-15.38\%}$} &
          {\ul $\textbf{-5.77\%}$} &
          {\ul $\textbf{-5.77\%}$} &
          {\ul $\textbf{-3.85\%}$} &
          $0.00\%$ &
          $0.00\%$ &
          $0.00\%$ &
          $0.00\%$ \\
        \bottomrule[1pt]
        \end{tabular}
	}
	}
	\caption{\label{tab:deltaAcc_leave_one}The relative difference of accuracy on each question type when only one head is kept in each layer, which is calculated by $(Acc_{new}-Acc_{org})/Acc_{org}$. In each layer, we only show the one with the \textbf{largest value} out of $12$ heads. For the result of all $12$ heads, please see Appendix B. Underlined numbers indicate that its absolute value is above $1\%$.
	}
	\end{table*}
\end{center}

\subsection{Automatic Dynamic Chopping}
In contrast to manual chopping, we hope the model can automatically learn how important and relevant a specific layer is. Therefore, we design a dynamic chopping module to explore the effect of each layer $l$ by learning a score $S_{l}$ for each layer, then remove layers with scores below a given threshold. An instance level layer score is calculated by:

\begin{equation}
{
	\begin{split}
    	& [\mathbf{q}^l_1, ..., \mathbf{q}^l_h, ..., \mathbf{q}^l_H] = \bW({hidden\_state^l}[0]), \\
	    & S_{l} = \mathbf{Sigmoid}(\frac{\textstyle{\sum_{i=1}^{H}} \mathbf{q}_{i}^l}{H}),
	\end{split}
	\label{layer_score}}
\end{equation}
where $\bW$ is a linear layer, $hidden\_state^l$ is the hidden\_state output by layer $l$ and we only take out the first element of this feature, which is also called \textbf{[CLS]} feature. $12$ head scores $\mathbf{q}^l_h$ are averaged and put into a $\mathbf{Sigmoid}$ function to yield a value between $0$ and $1$, measuring the degree of importance for a specific layer. The score matrices will change when dealing with different instances.

Then we use this layer score $S_{l}$ to determine whether to chop a layer or not. If $S_{l}$ is below a threshold, then the Layer $l$ will be chopped at the forward time. To chop the Layer $l$, we directly skip it and connect the output of Layer $l-1$ to the input of the Layer $l+1$. 

The dynamic chopping module is trained independently without modifying any parameters in the Transformer model. The maximal learning rate is $1e-3$ and the batch size is $480$. This module is trained with a binary cross-entropy loss (for answer prediction) and an additional loss ($L1$ norm of head scores). 

\begin{figure*}[t!]
	\begin{center}
		\includegraphics[width=0.98\textwidth]{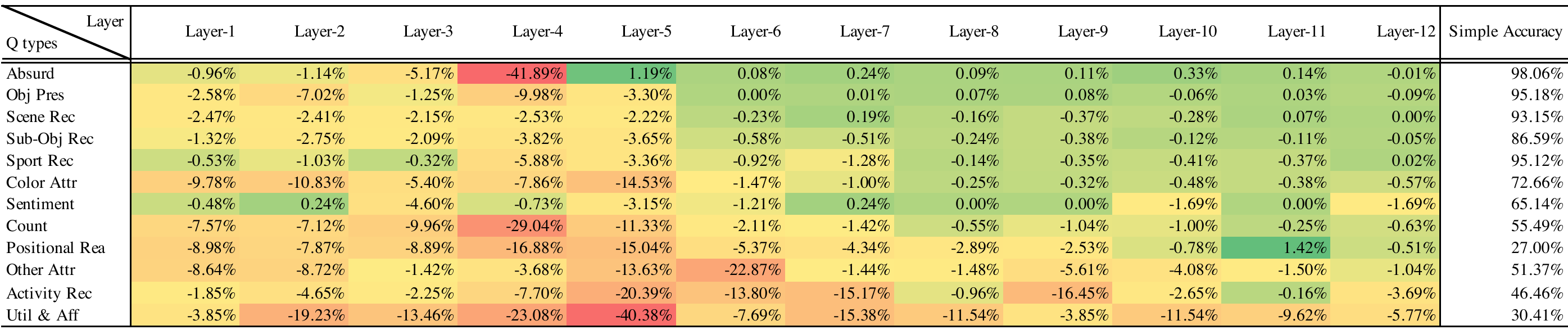}
	\end{center}
	\caption{The relative difference of accuracy on each type when one layer (all $12$ heads) is removed, which is calculated by $(Acc_{new}-Acc_{org})/Acc_{org}$. The matrix shows an interesting echelon form, which reveals the tendency of Transformer to rely on higher-level layers for higher-level questions.}
	\label{fig:rm_one_layer}
\end{figure*}

\begin{figure*}[t!]
	\begin{center}
		\includegraphics[width=0.98\textwidth]{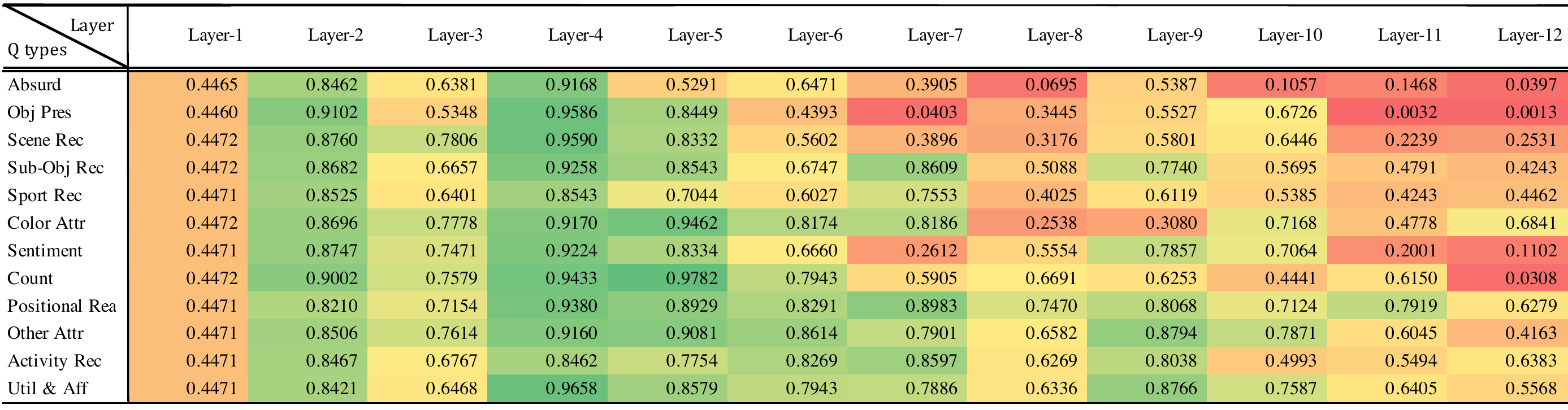}
	\end{center}
	\caption{Instance level layer score (an average of all instances) in each question type learned by attention on \textbf{[CLS]} feature. The result matrix here is also in an echelon form, similar to that of Figure \ref{fig:rm_one_layer}.}
	\vspace{-0.3cm}
	\label{fig:learned_head_score}
\end{figure*}

\section{Experiments}
\subsection{Experimental Settings}
We thoroughly study the effect of Transformer's heads and layers on the VQA task when questions are in different types. All of our experiments are conducted on the Task Driven Image Understanding Challenge (TDIUC)~\cite{kafle2017analysis} dataset, a large VQA dataset. This dataset was proposed to compensate for the bias in distribution of different question types of VQA 2.0~\cite{goyal2017making}. It is larger than VQA 2.0 and divides questions into $12$ more fine-grained categories based on the task they solve, which provide us convenience to 
do analysis by question types.

TDIUC has $12$ question types, including both classical computer vision tasks and novel high-level vision tasks which require varying degrees of image understanding and reasoning. The question types are: \texttt{Absurd}, \texttt{Object Presence}, \texttt{Scene Recognition}, \texttt{Subordinate Object Recognition}, \texttt{Sport Recognition}, \texttt{Color Attributes}, \texttt{Sentiment Understanding}, \texttt{Counting}, \texttt{Positional Reasoning}, \texttt{Other Attributes}, \texttt{Activity Recognition}, and \texttt{Object Utilities $\&$ Affordances}. The total number of unique answers is $1,618$, which is then set as the size of our answer vocabulary. 

Experiments are conducted on 4 NVIDIA GeForce 2080Ti GPUs with a batch size of $480$. We choose a general and popular pre-training model VisualBERT~\cite{li2019visualbert}, which is built on top of PyTorch. This model consists of visual embeddings $\&$ text embeddings, a one-stream Transformer and a classifier specific for VQA tasks. There are $L=12$ layers of encoder, and each layer has $H=12$ heads. The hidden state dimension is $768$.
We load the model pre-trained on COCO Caption~\cite{chen2015microsoft} dataset, then finetune it with a leaning rate of $5e-5$ on the TDIUC dataset. 
The image features are extracted from a ResNet-101~\cite{he2016deep} based Faster R-CNN~\cite{ren2015faster} model pre-trained on Visual Genome~\cite{krishna2017visual}.

\subsection{Effect of Manual Chopping}
\label{sec:effect_manual_chopping}
\subsubsection{Removing One Head at One Time}
First, we follow the previous work~\cite{michel2019sixteen} to study if there is a specific role for a particular attention head $h$ in different question types. To study the contribution of a particular head, we mask a single head (\ie multiplying $\textbf{Att}_h^l$ with zero-valued binary mask) in Transformer at a time, then evaluate the model's performance. If one head is important in a certain question type, the accuracy after the removal of this head will decrease dramatically. On the other hand, the increase in accuracy after removal can tell that this head is insignificant or even plays a negative role in this question type. 

In Table \ref{tab:deltaAcc_removee_one}, we show the relative difference of accuracy ($(Acc_{new}-Acc_{org})/Acc_{org}$) on each type when only one head is removed. In each layer, there are $12$ heads and we show the one who has the \textbf{largest absolute value} of relative difference to get a clue about the maximal influence of one layer. In most of the question types, the largest absolute value of relative difference out of $12$ heads is still a small number, meaning some heads in this layer can be removed to get a more efficient model. In question types such as ``\texttt{Activity Recognition}'' and ``\texttt{Utilities $\&$ Affordances}'', removal of a specific head in more layers can bring a dramatic decrease in accuracy, which can illustrate that more layers (both low-level and high-level) are needed here. Observing the percentage of change in accuracy for each head (can be found in Appendix A), we found that in the majority of question types, removing one head does not hurt the accuracy a lot. Moreover, removal of some heads can even bring the improvement of accuracy, which means heads are not equally important in answering a specific question type. 

\begin{center}
    \begin{table*}[t!]
    \centering
    \scalebox{0.7}{
    \setlength{\tabcolsep}{0.5mm}{
    \begin{tabular}{llccc|ccccccc}
    \toprule[1pt]
    \multicolumn{1}{c}{\#} &
      \multicolumn{1}{c}{Question types} &
      \begin{tabular}[c]{@{}c@{}}NMN\\ {\cite{andreas2015deep}}\end{tabular} &
      \begin{tabular}[c]{@{}c@{}}RUA\\ {\cite{noh2016training}}\end{tabular} &
      \begin{tabular}[c]{@{}c@{}}QTA\\ {\cite{shi2018question}}\end{tabular} &
      \begin{tabular}[c]{@{}c@{}}VisualBERT\\ (random 50\%)\end{tabular} &
      \begin{tabular}[c]{@{}c@{}}VisualBERT\\ (\textgreater{}0.05)\end{tabular} &
      \begin{tabular}[c]{@{}c@{}}VisualBERT\\ (\textgreater{}0.1)\end{tabular} &
      \begin{tabular}[c]{@{}c@{}}VisualBERT\\ (\textgreater{}0.3)\end{tabular} &
      \begin{tabular}[c]{@{}c@{}}VisualBERT\\ (\textgreater{}0.5)\end{tabular} &
      \begin{tabular}[c]{@{}c@{}}VisualBERT\\ (\textgreater{}0.7)\end{tabular} &
      \begin{tabular}[c]{@{}c@{}}VisualBERT\\ (full)\end{tabular} \\ \hline\hline
    1  & Absurd          & $87.51$ & $96.08$ & $100.0$ & $4.02$  & $98.07$          & $98.07$          & $98.10$          & $\textbf{98.21}$ & $84.40$ & $98.06$          \\
    2  & Obj Pres        & $92.50$ & $94.38$ & $94.55$ & $52.01$ & $95.19$          & $95.25$          & $\textbf{95.32}$ & $94.52$          & $72.46$ & $95.20$          \\
    3  & Scene Rec       & $91.88$ & $93.96$ & $93.80$ & $34.68$ & $93.15$          & $93.18$         & $\textbf{93.36}$ & $93.05$          & $73.73$ & $93.15$          \\
    4  & Sub-Obj Rec     & $82.02$ & $86.11$ & $86.98$ & $4.07$  & $86.59$          & $\textbf{86.62}$ & $86.61$          & $86.38$          & $70.14$ & $86.59$          \\
    5  & Sport Rec       & $89.99$ & $93.47$ & $95.55$ & $14.91$ & $95.12$          & $\textbf{95.13}$ & $\textbf{95.13}$ & $94.22$          & $67.66$ & $95.12$          \\
    6  & Color Attr      & $54.91$ & $66.68$ & $60.16$ & $21.83$ & $72.62$          & $\textbf{72.66}$ & $72.61$          & $71.05$          & $64.20$ & $\textbf{72.66}$          \\
    7  & Sentiment       & $58.04$ & $60.09$ & $64.38$ & $39.27$ & $65.14$          & $64.98$          & $\textbf{65.46}$ & $64.67$          & $50.16$ & $65.14$          \\
    8  & Count           & $49.21$ & $48.43$ & $53.25$ & $20.52$ & $55.40$          & $55.39$          & $55.46$          & $53.82$          & $42.05$ & $\textbf{55.49}$ \\
    9  & Positional Rec  & $27.92$ & $35.26$ & $34.71$ & $0.90$  & $27.02$          & $\textbf{27.04}$ & $26.95$          & $26.03$          & $19.36$ & $27.00$          \\
    10 & Other Attr      & $47.66$ & $56.49$ & $54.36$ & $1.59$  & $\textbf{51.37}$ & $\textbf{51.37}$ & $51.28$          & $50.38$          & $47.13$ & $\textbf{51.37}$ \\
    11 & Activity Rec    & $44.26$ & $51.60$ & $60.10$ & $0.00$  & $46.56$          & $46.46$          & $\textbf{46.61}$ & $43.73$          & $25.09$ & $46.46$          \\
    12 & Util \& Aff     & $25.15$ & $31.58$ & $31.48$ & $10.53$ & $30.41$          & $\textbf{30.99}$ & $\textbf{30.99}$ & $27.49$          & $22.22$ & $30.41$          \\ \hline
    13 & A-MPT           & $62.59$ & $67.81$ & $69.11$ & $42.54$ & $68.04$          & $68.10$          & $\textbf{68.16}$ & $66.96$          & $53.22$ & $68.05$          \\
    14 & H-MPT           & $51.87$ & $59.00$ & $60.08$ & $33.61$ & $56.72$          & $56.89$          & $\textbf{56.91}$ & $54.71$          & $42.07$ & $56.73$          \\
    15 & Simple Accuracy & $79.56$ & $84.26$ & $87.52$ & $28.25$ & $86.13$          & $86.16$          & $\textbf{86.21}$ & $85.47$          & $69.10$ & $86.51$          \\ 
    \bottomrule[1pt]
    \end{tabular}
    }
    }
    \vspace{-0.1cm}
    \caption{The accuracy on TDIUC dataset after removing Transformer layers whose scores are below the specific thresholds. The highest accuracy values on the right part of the table are in bold. When we remove layers whose scores are below $0.3$, the accuracy, arithmetic mean-per-type (A-MPT) accuracy and harmonic mean-per-type accuracy (H-MPT) all surpass those of the full VisualBERT model, and over half of the question types also increase in accuracy. }
    \vspace{-0.2cm}
    \label{tab:acc_threshold}
    \end{table*}
\end{center}

\begin{figure}[t!]
	\begin{center}
		\includegraphics[width=0.48\textwidth]{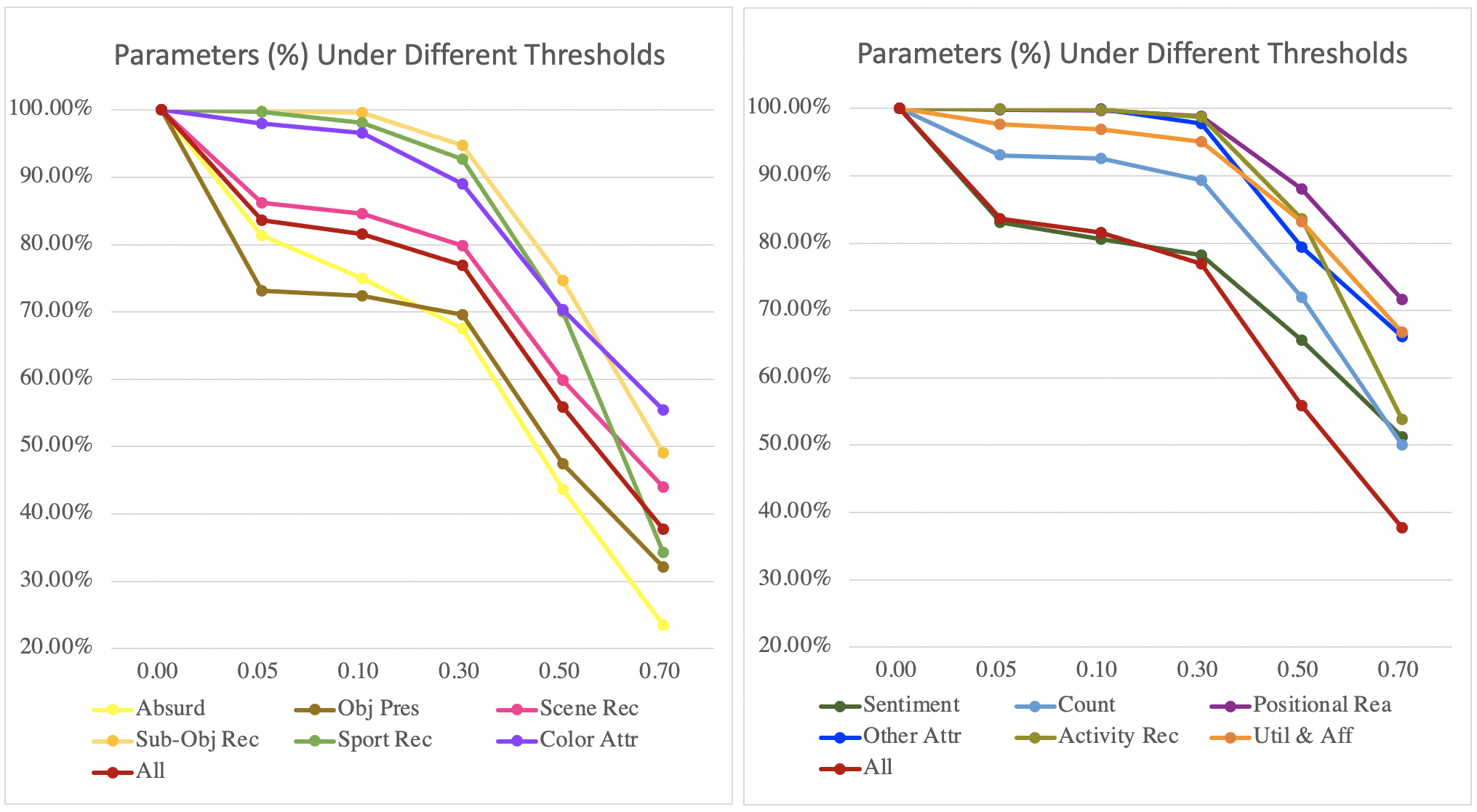}
	\end{center}
	\vspace{-8pt}
	\caption{The chopping curve, showing percentage of Transformer parameters after chopping layers under the control of several different thresholds.}
	\vspace{-0.2cm}
	\label{fig:params}
\end{figure}

\vspace{-23pt}
\subsubsection{Removing All Heads but One}
It is also a question whether more than one head $h$ is needed in a specific layer. We mask all heads in a layer except for a single one and show the relative difference of accuracy on each question type in Table \ref{tab:deltaAcc_leave_one}. In each layer, we only show the \textbf{largest value} of relative difference out of $12$ heads ($\mathbf{max}((Acc^i_{new}-Acc^i_{org})/Acc^i_{org}), i \in [1,12]$). If there exists one head that can replace or even surpass all heads in a layer, this largest value will be a positive value. For some layers in a specific question type, one head is indeed sufficient at test time. 

On the other hand, in some layers even the largest value of relative difference is a negative number, which means the high accuracy comes from the cooperative work of multiple heads. In question types \texttt{Color Attributes} and \texttt{Activity Recognition}, nearly none of a single head can replace a whole layer in all $12$ layers. 
Besides, observing the relative difference of accuracy for each head (in Appendix B), we find that most of the heads in one layer tend to have the same level of influence. To further investigate whether there are some specific roles for different layers, we conduct experiments by removing a whole layer at a time below. 

\begin{figure*}[t!]
	\begin{center}
		\includegraphics[width=0.85\textwidth]{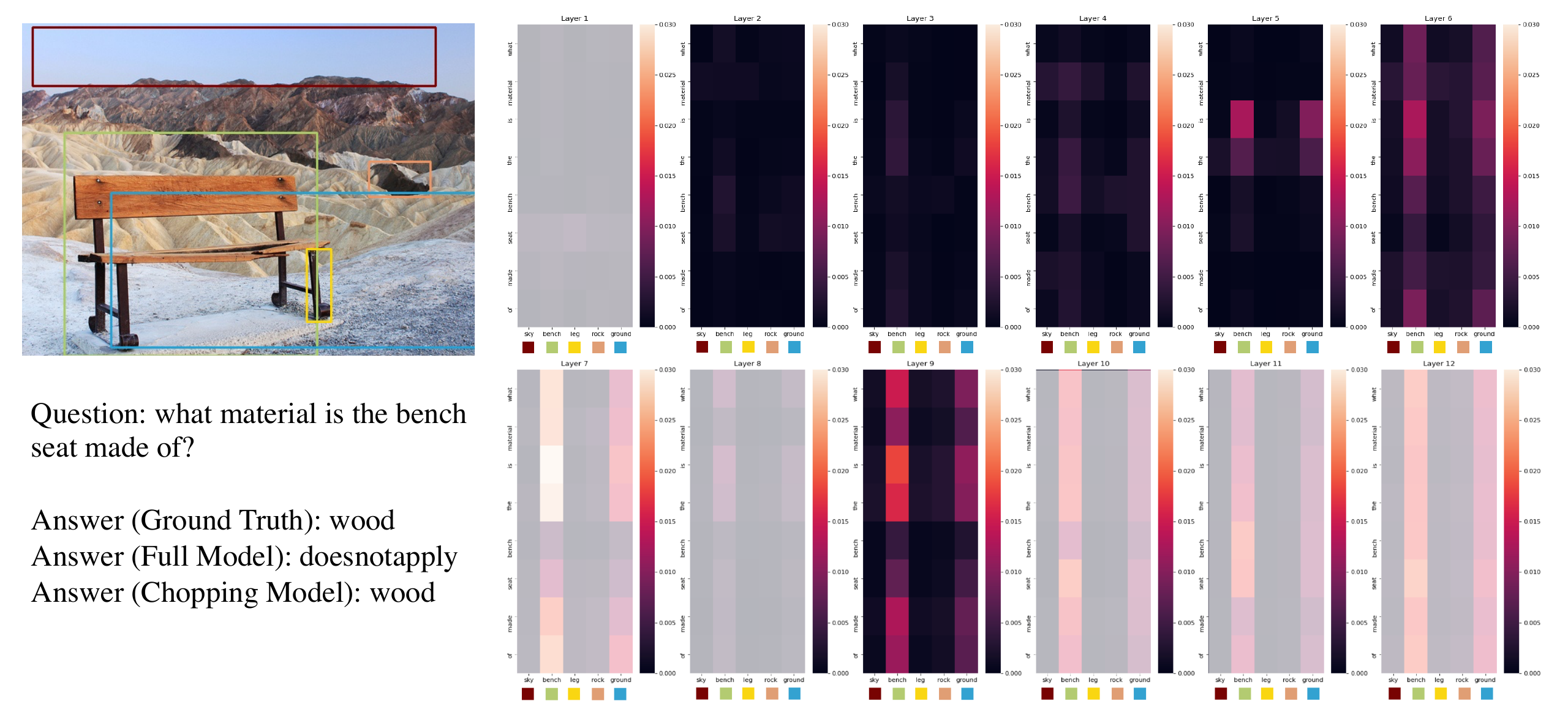}
	\end{center}
	\vspace{-0.5cm}
	\caption{Attention weights of each layers in VisualBERT. Here we only show the attention weights matrices of the question and top-5 objects. The weights shown are the average of a layer (all $12$ heads). Layers covered by semi-transparent rectangular are chopped under the threshold ($0.5$). }
	\vspace{-0.2cm}
	\label{fig:vis_attn}
\end{figure*}

\subsubsection{Removing One Layer}
To investigate the effect of layers, we remove 
one layer in Transformer at a time. 
The dramatically decrease of accuracy illustrates that a layer plays a vital role in a question type. The more the accuracy decrease, the more important a layer is. While a positive relative difference tells that a layer is useless and can be removed. From the results in Figure \ref{fig:rm_one_layer}, removing most of the layers cause a more or less decrease of accuracy, while other layers bring a slight increase of accuracy.

Results show that lower-level layers are enough for simple questions, such as questions belong to \texttt{Absurd}, \texttt{Object Presence}\etc; higher-level layers are important in questions where more reasoning abilities are needed such as questions belong to \texttt{Sentiment Understanding}, \texttt{Activity Recognition}\etc The $12 \times 12$ result matrix in Figure \ref{fig:rm_one_layer} also shows an interesting echelon shape that gradually changes according the level of reasoning. Questions belonging to \texttt{Other Attributes} (\eg{‘What shape is the clock?’}) and \texttt{Object Utilities and Affordances} (\eg{‘What object can be used to break glass?’}) may ask the materials or shapes of objects, which needs strong reasoning ability and common sense knowledge to answer them; \texttt{Activity Recognition} (\eg{‘What is the girl doing?’}) requires scene and activity understanding abilities; \texttt{Positional Reasoning} (\eg{‘What is to the left of the man on the sofa?’}) and \texttt{Sentiment Understanding} (\eg{‘How is she feeling?’}) are two kinds of well known complicated questions requiring multiple reasoning 
and understanding abilities. Removing high-level layers will dramatically hurt accuracy in the above-mentioned question types. 

We also observe some other interesting phenomena. In question type \texttt{Scene Recognition} (\eg{'What room is this?'}) removal of the first $5$ layers causes decrease of accuracy in roughly a same manner, reflecting that they are all important for scene understanding. In question type \texttt{Object Presence} (\eg{‘Is there a cat in the image?’}) Layer $4$ plays the most important role. Layer $4$ is also important in other questions classes related to objects such as \texttt{Subordinate Object Recognition}, \texttt{Counting} (where the model needs to count the number of objects) and \texttt{Color Attributes} (locate the object firstly then recognize its color). The question type \texttt{Absurd} (\ie Nonsensical queries about the image) is the easiest class, and it can be answered easily by determining whether the object mentioned in the question appears in the image, which also depends on objects and the removal of Layer $4$ impacts on the accuracy greatly. Layer $3$ and Layer $5$ play a critical role in question type \texttt{Counting}; Layer $2$ and Layer $5$ are important in question type \texttt{Color Attributes}, which illustrates that some layers cooperate to work out special reasoning problems.

\subsection{Effect of Dynamic Chopping}
\label{sec:layre_chopping}
In our previous section, we analyze the effect of ‘removing one head’, ‘removing all heads but one within a single layer’ and ‘directly removing one layer’. Based on the observations, we consider a single layer as a basic module in Transformer and want to further explore what would happen if we chop two or more layers at the same time. In the dynamic chopping module, a score is learned for each layer and we set several thresholds to determine whether to remove a layer or not, then we record the change in accuracy.  

The mean of layer scores is collected in each type and the results can be found in Figure \ref{fig:learned_head_score}. The distribution of layer scores is similar to that of relative difference in manually removing one layer (Figure \ref{fig:rm_one_layer}). When removing a layer hurts the accuracy a lot, the dynamic chopping module tends to learn a higher score for this layer. 
The layer scores matrix here reflects importance of multiple layers simultaneously, while only one layer can be tested in manually removing one layer.

We set several thresholds to find the best trade-off between accuracy and model size. Layers whose score is lower than the threshold value are removed. All the results can be found in Table \ref{tab:acc_threshold}. 
The percentage of parameters under the control of different thresholds are shown in Figure \ref{fig:params} and the detailed value of parameters can be found in Appendix D. For lower-level questions, the chopping curve is steeper, meaning it is easier to chop more layers. However, for higher-level questions, the chopping curve is smoother, as the chopping of layers needs stricter thresholds.

With larger thresholds, fewer parameters are kept by the VisualBERT model. When we set threshold to ($0.3$), the model achieves higher accuracy on half of the question types with roughly $76.92\%$ of parameters reserved. When we reserve layers whose score is larger than $0.5$, only $55.85\%$ of parameters are reserved, with only $1\%$ loss of accuracy. When we set threshold to ($0.7$), more parameters will be removed, with a significant impact on performance. However, to remove $50\%$ layers randomly yields a much worse performance. The results show that our dynamic chopping module successfully predict scores to control which layer can be chopped efficiently. 
We also observe that one layer cannot replace the whole Transformer (Appendix C), and the result shows leaving only one layer in almost all question types cause over $50\%$ of accuracy drop. All the results illustrate that not all layers can be reduced without significantly influencing performance. 
A qualitative example is also shown in Figure \ref{fig:vis_attn}. After chopping insignificant layers under the threshold ($0.5$), the surviving layers can generate the true answer. Please refer to Appendix E and Appendix F for more examples and the analysis about failure cases.

\vspace{-0.1cm}
\section{Conclusion}
In this paper, we conduct extensive experiments on a Visual Question Answering dataset categorized by question types and explore the roles of heads and layers in a Vision-and-Language pre-trained Transformer by ablating these heads and layers separately. Interesting results as shown in an echelon shape reveal higher-level layers are required for higher-level questions. A dynamic chopping module is further designed to automatically learn efficiently chopped model. We hope our findings to advance the understanding of Transformer structure under different level of visual reasoning question types and inspire the investigation of more accurate and more efficient visual reasoning models.

\bibliographystyle{named}
\bibliography{ijcai21}

\onecolumn
\newpage

\newcommand{\xiaoer}{\fontsize{14pt}{14pt}\selectfont}
\begin{center}
\xiaoer{\textbf{Appendix}}
\end{center}
\setcounter{section}{0}
\renewcommand\thesection{\Alph{section}} 

\section{Removing One Head}
\begin{figure*}[b!]

	\begin{center}
		\includegraphics[width=0.84\textwidth]{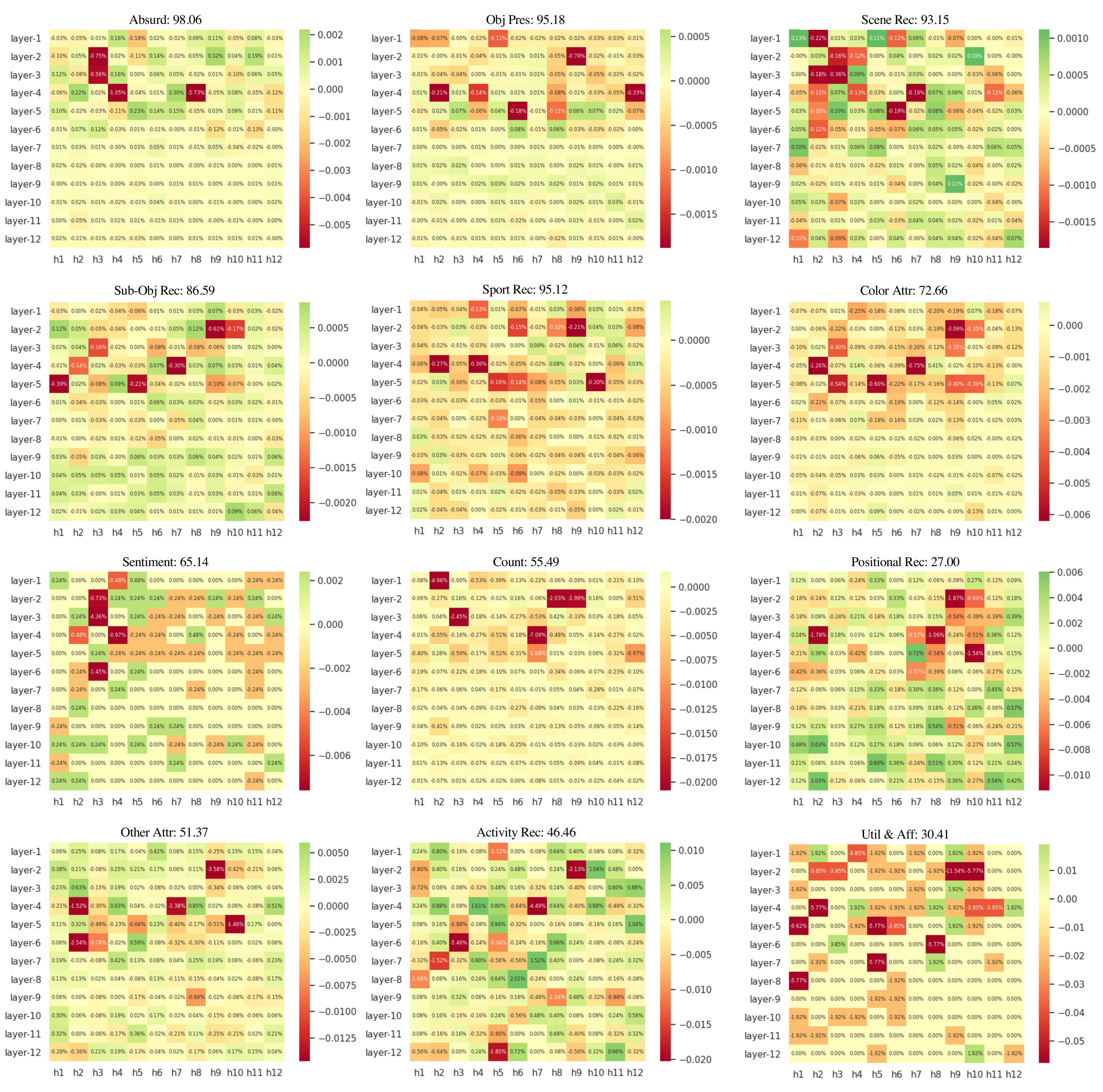}
	\end{center}
	\caption{The relative difference of accuracy on each category when one head is removed, calculated by $(Acc_{new}-Acc_{org})/Acc_{org}$.}
	\label{fig:supp_remove_one}
\end{figure*}

Figure\ref{fig:supp_remove_one} shows the relative difference of accuracy on each category when only one head is removed, illustrated with results of all heads for complete reference. In the majority of question classes, removing one head does not hurt the accuracy a lot, and sometimes can even bring the improvement of accuracy. 

\newpage
\section{Removing All Heads but One}
\begin{figure*}[b!]
    \vspace{-5cm}
	\begin{center}
		\includegraphics[width=0.98\textwidth]{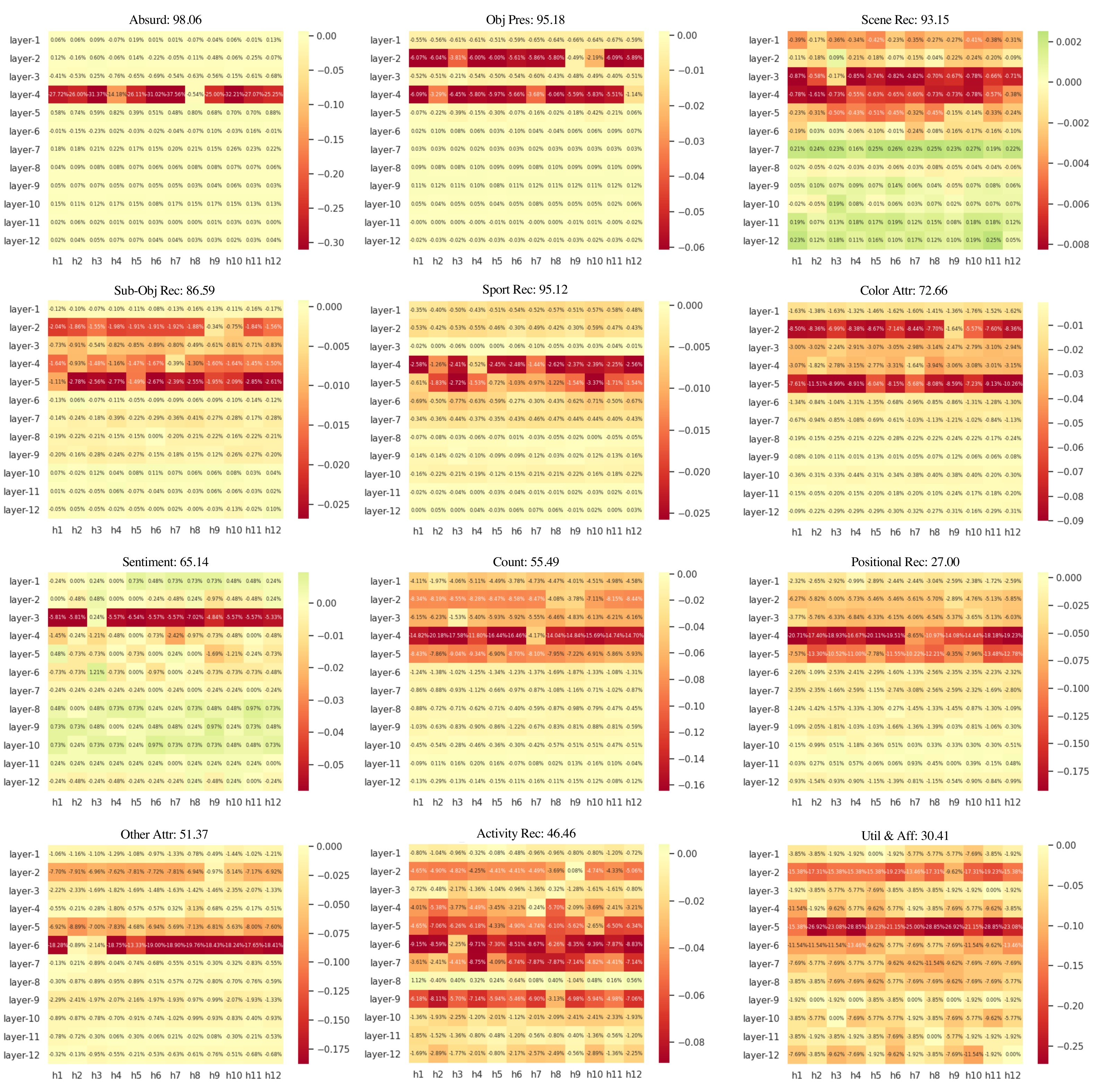}
	\end{center}
	\caption{The relative difference of accuracy on each category when all heads but one in a layer are removed, which is calculated by $(Acc_{new}-Acc_{org})/Acc_{org}$.}
	\vspace{2cm}
	\label{fig:supp_leave_one}
\end{figure*}

Figure\ref{fig:supp_leave_one} shows the relative difference of accuracy on each category when all heads but one in a layer are removed, illustrated with results of all heads for complete reference. Most of the heads in the same layer are inclined to have the same level of influence. 

\newpage
\section{Removing All Layers but One}
\begin{figure*}[t!]
	\begin{center}
		\includegraphics[width=0.98\textwidth]{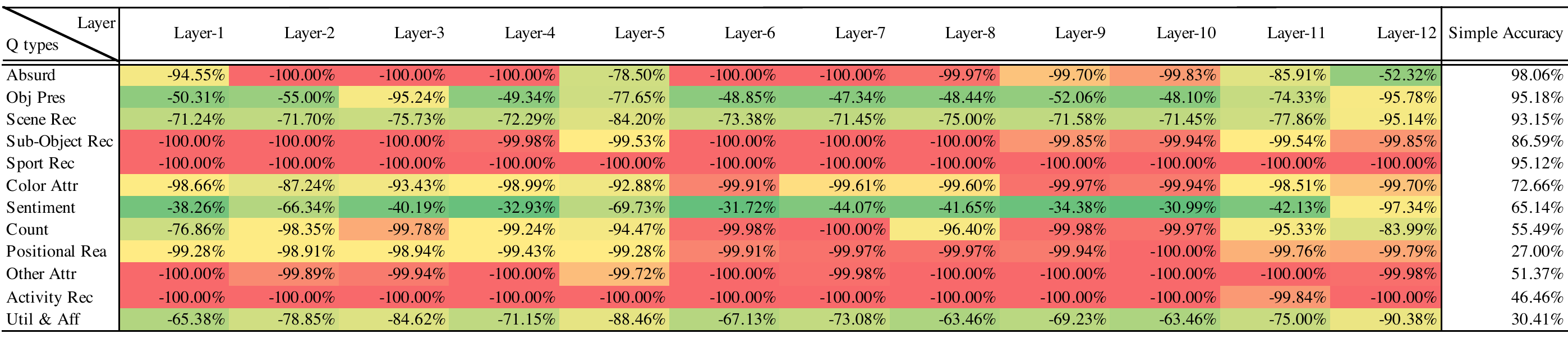}
	\end{center}
	\caption{The relative difference of accuracy on each category when removing all layers but one, calculated by $(Acc_{new}-Acc_{org})/Acc_{org}$.}
	\label{fig:supp_rm_layer}
\end{figure*}

Figure\ref{fig:supp_rm_layer} shows the relative difference of accuracy on each category when all layers ($12$ layers) but one are removed. It shows that leaving only one layer in almost all question categories causes over $50\%$ of accuracy drop, illustrating that only one layer is not sufficient enough to replace the whole Transformer in visual question answering tasks.

\section{Parameters Under Different Thresholds }
\begin{figure*}[t!]
	\begin{center}
		\includegraphics[width=0.98\textwidth]{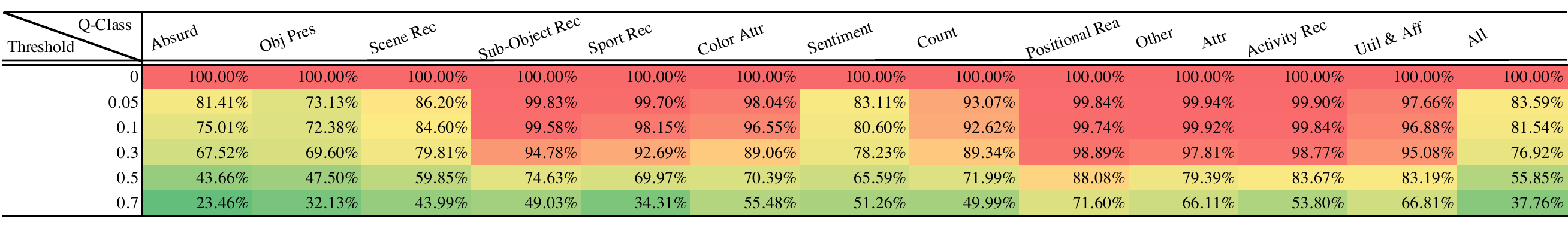}
	\end{center}
	\caption{The precise value of parameters remained in Transformer after chopping layers under the control of different thresholds.}
	\label{fig:supp_param}
\end{figure*}

Figure\ref{fig:supp_param} shows the precise value of parameters remained in Transformer after chopping layers under the control of different thresholds.

\section{Additional Success Cases}
\begin{figure*}[b!]
	\begin{center}
		\includegraphics[width=0.98\textwidth]{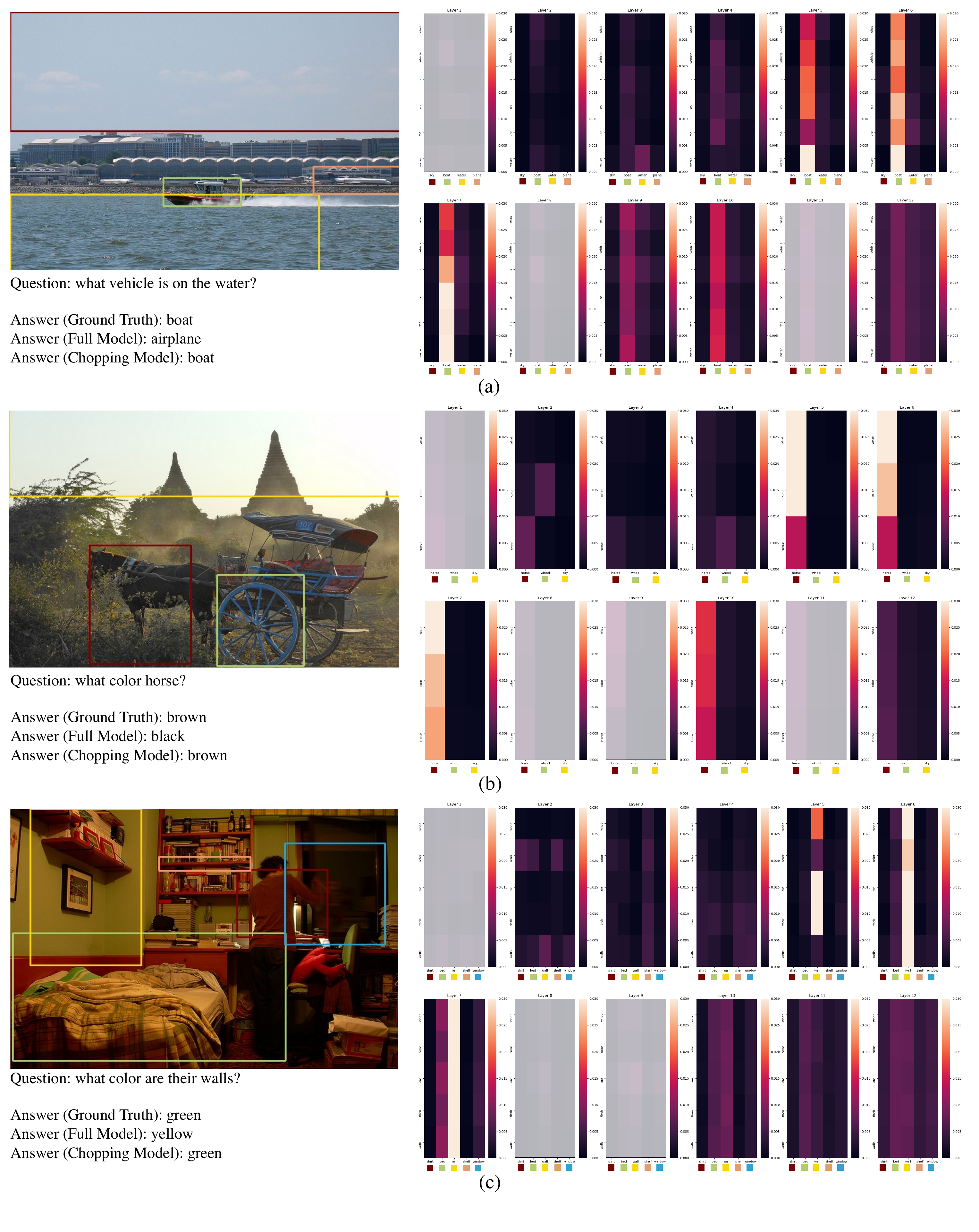}
	\end{center}
	\caption{Additional qualitative examples. After chopping some insignificant or useless layers, the surviving layers can generate the correct answer. Layers covered by semi-transparent rectangular are chopped under the control of threshold ($0.5$). }
	\label{fig:supp_succ_choppingTrue}
\end{figure*}

In Figure\ref{fig:supp_succ_choppingTrue}, we show the additional examples that after chopping some insignificant layers, the remaining layers can generate the correct answer. 

\begin{figure*}[b!]
	\begin{center}
		\includegraphics[width=0.98\textwidth]{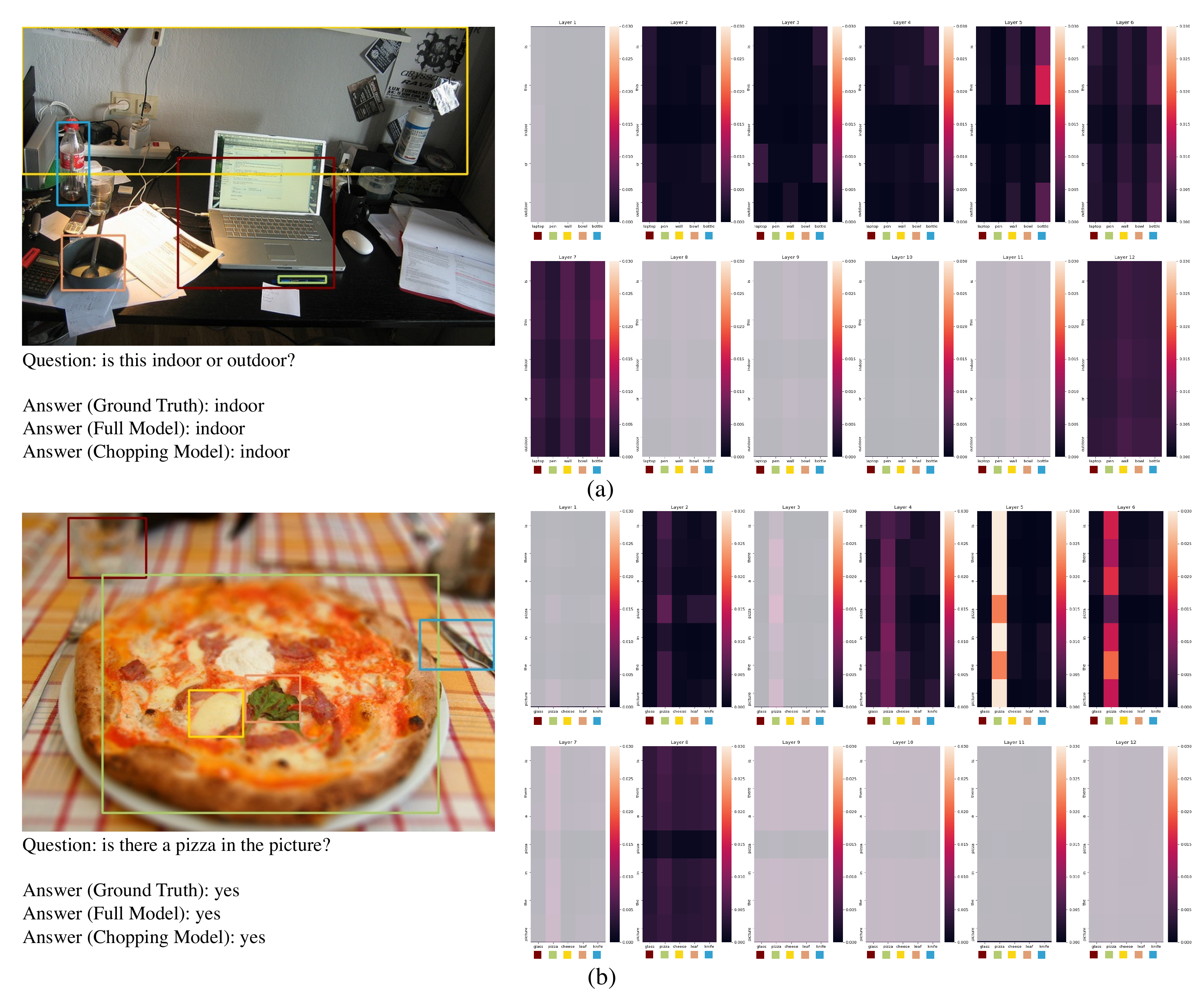}
	\end{center}
	\caption{Both the full model and our chopped model can predict the correct answer. Interestingly, after chopping some insignificant layers, the model still has the ability to predict the correct answer. Layers covered by translucent rectangular are chopped under the control of threshold ($0.5$). }
	\label{fig:supp_succ_allTrue}
\end{figure*}

In Figure\ref{fig:supp_succ_allTrue}, both the full model and our chopped model can predict the correct answer. A model can determine whether the scene is "indoor" or "outdoor" in (a) by recognizing some objects with specific characteristics, \eg "laptop", "bottle" and "wall" always jointly appear in indoor scenes. To answer "is there a pizza" also mainly depends on the recognition of objects. All questions above can be answered correctly with fewer parameters by chopping some insignificant layers.

\section{Failure Cases}
\begin{figure*}[t!]
	\begin{center}
		\includegraphics[width=0.98\textwidth]{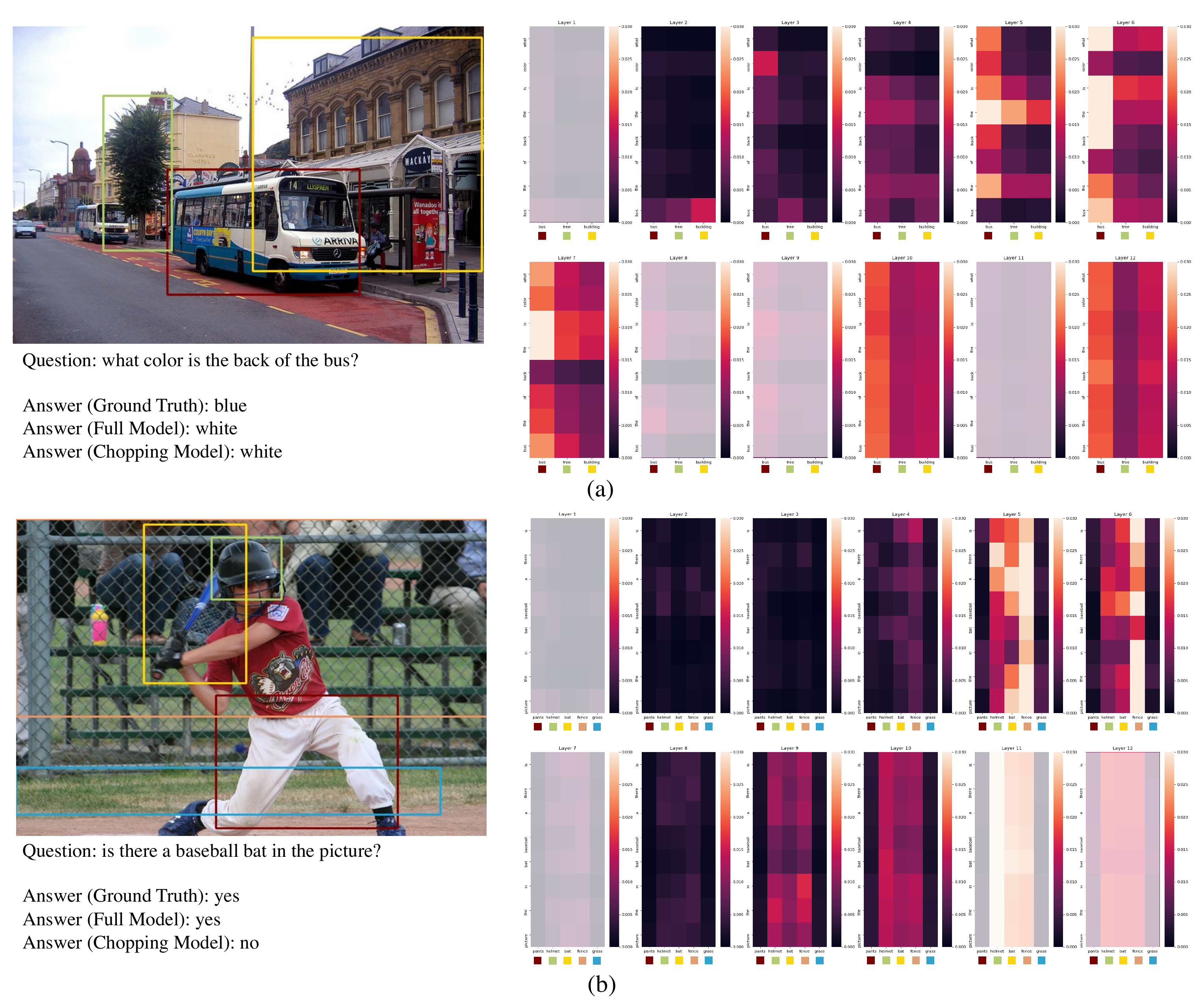}
	\end{center}
	\caption{Examples where chopped model fails. Layers covered by semi-transparent rectangular are chopped under the control of threshold ($0.5$). }
	\label{fig:supp_failure_chopping}
\end{figure*}

Our chopped model fails to answer the questions in Figure\ref{fig:supp_failure_chopping}. In Figure\ref{fig:supp_failure_chopping} (a), the question asks about "back color" of a bus, a more fine-grained characteristic, requiring more and deeper layers of Transformer to reason jointly. Similarly, question in (b) asks about the appearance of "baseball bat" while the detector can only detect a "bat". To determine what sports this "bat" is for, a comprehensive analysis of all parts of the entire scene is needed. 

\begin{figure*}[t!]
	\begin{center}
		\includegraphics[width=0.98\textwidth]{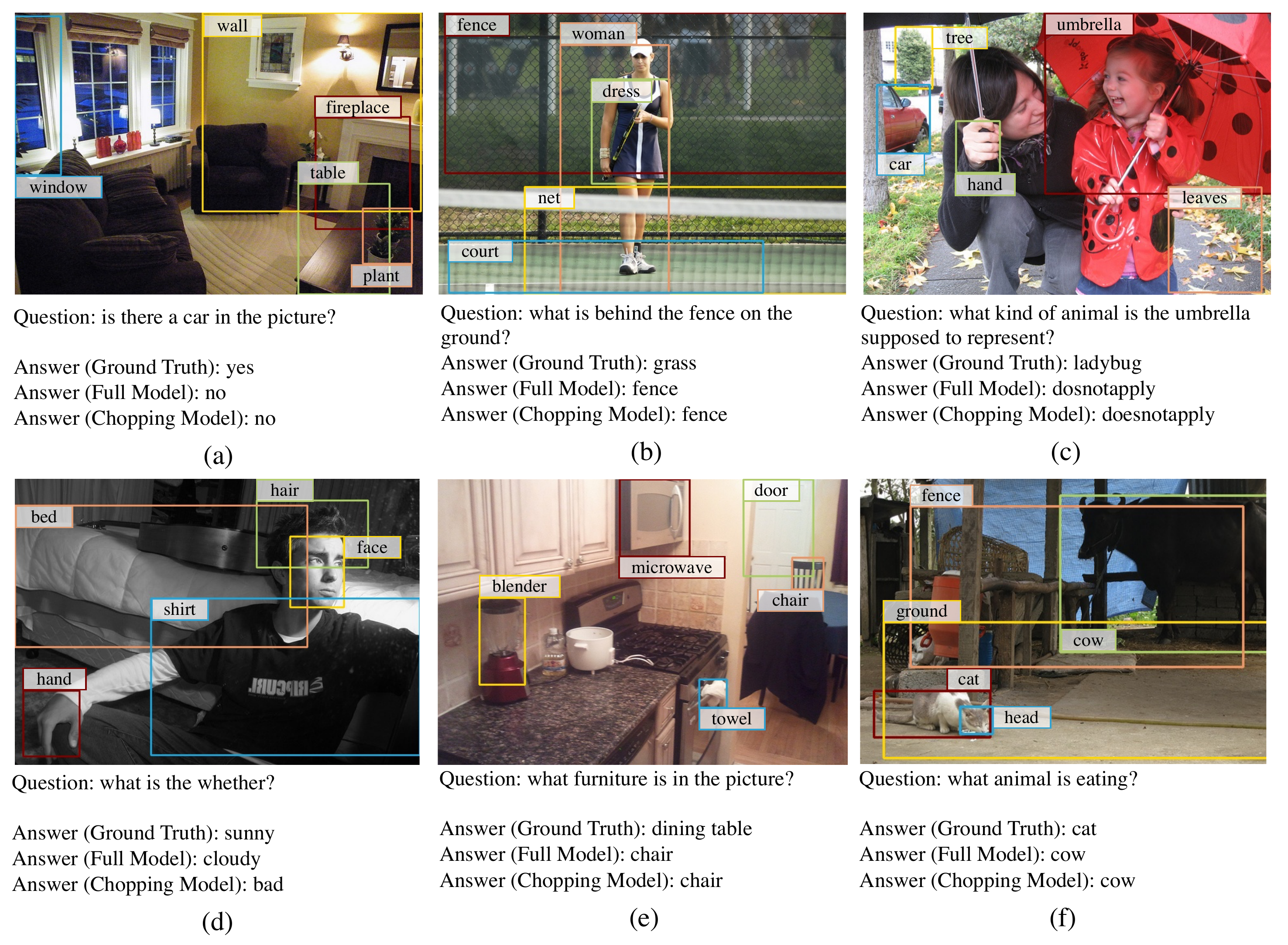}
	\end{center}
	\caption{Examples both full model and chopping model fail to answer. }
	\label{fig:supp_failure_question}
\end{figure*}

Other questions that our model fails to answer can be categorized into several groups. Firstly, the objects most related to question are not detected. In the Figure\ref{fig:supp_failure_question}, question (a) and (b) both ask about some objects hard to detect, such as the car in (a) which is even hard for human to find (out of a bright room's window at night) and grass in (b) all appears behind the fence while our detector can only detect the fence in the foreground. Secondly, some questions need strong reasoning abilities and commonsense knowledge even imagination abilities to answer "the umbrella supposed to represent which animal" in (c). Thirdly, the question is hard to answer in (d) as it hard to determine whether it is a sunny day from a black and white photo. Finally, some questions have multiple true answers while each question in TDIUC only have one ground truth answer so our model fails in (e) and (f). 

\end{document}